\documentclass[letterpaper]{article}
\usepackage[utf8]{inputenc}
\usepackage[T1]{fontenc}
\usepackage{aaai}
\usepackage{times}
\usepackage{courier}
\usepackage{amsmath,amssymb,amsthm}
\usepackage{mathtools}
\usepackage{newtxmath}
\usepackage[authoryear,square]{natbib}
\usepackage{array}
\usepackage{booktabs}
\usepackage{cleveref}
\usepackage{microtype}

\usepackage{geometry}
\usepackage{setspace}
\usepackage{enumerate}
\usepackage{psfrag}

\title{Complex and Holographic Embeddings of Knowledge Graphs: \\A Comparison}

\author{%
  Th\'eo Trouillon\\
  Univ. Grenoble Alpes\thanks{Work also done while at Xerox Research Centre Europe.}\\
  \texttt{theo.trouillon@gmail.com}
  \And
  Maximilian Nickel\\
  Facebook AI Research\\
  Massachusetts Institute of Technology, LCSL\\
  \texttt{maxn@fb.com}
}

\newtheorem{thm}{Theorem}

\renewcommand{\cite}{\citep}

\newcommand{\complexSpace}{\mathbb{C}}
\renewcommand{\Re}{\mathbb{R}}
\newcommand{\C}{\complexSpace} 
\newcommand{\R}{\Re} 
\newcommand{\rk}{K}

\newcommand{\real}{\mathrm{Re}}

\newcommand{\T}{^\top}
\newcommand{\conj}{\overline}

\newcommand{\Ne}{N_e} 
\newcommand{\Nr}{N_r} 
\newcommand{\setobs}{\mathcal{D}} 
\newcommand{\setent}{\mathcal{E}} 
\newcommand{\setrel}{\mathcal{R}} 

\newcommand{\score}{\phi}
\newcommand{\ev}{e} 
\newcommand{\Em}{E} 
\newcommand{\rv}{r} 
\newcommand{\Rm}{R} 

\newcommand{\F}{\mathcal{F}}

\usepackage[textsize=small]{todonotes}

\graphicspath{{figures/}}

\begin{document}
\maketitle

\section{Introduction}
Embeddings of knowledge graphs have received significant attention due to their
excellent performance for tasks like link prediction and entity resolution. In
this short paper, we are providing a comparison of two state-of-the-art
knowledge graph embeddings for which their equivalence has recently been established, i.e., 
\textsc{ComplEx} and \textsc{HolE} \citep{nickel_2016_holographic,trouillon2016,hayashi2017equivalence}. 
First, we briefly review both models and discuss how their scoring functions are equivalent. 
We then analyze the discrepancy of results reported in the original 
articles, and show experimentally that they are likely due to the use
of different loss functions. In further experiments, we evaluate the ability of
both models to embed symmetric and antisymmetric patterns. Finally, we discuss
advantages and disadvantages of both models and under which conditions one would be preferable to the other.


\section{Equivalence of Complex and Holographic Embeddings}
\label{sec:equiv}
In this section, we will briefly review Holographic and Complex
embeddings and discuss the
equivalence of their scoring functions.

Let $\mathcal{G} = (\setent, \setrel, \mathcal{T})$ be a knowledge graph, which
consists of entities $\setent$, relation types $\setrel$ and observed triples
$\mathcal{T} \subseteq \setrel \times \setent \times \setent$.
Furthermore, let $\setobs$ be a training set, which associates with each
\emph{possible} triple in $\mathcal{G}$ its truth values $y \in \{\pm1\}$. That
is, for a possible triple $(p,s,o)$ 
with $s,o \in \setent$ and $p \in \setrel$ it holds that
\begin{equation*}
  y_{pso} =
  \begin{cases}
    +1, & \text{if } (p,s,o) \in \mathcal{T} \\
    -1, & \text{otherwise.}
  \end{cases}
\end{equation*}
For knowledge graphs with a large number of possible triples we employ negative
sampling as proposed by \citet{bordes2013translating}.
The objective of knowledge graph completion is then to learn a scoring function 
${\score_{pso} : \setrel \times \setent \times \setent \rightarrow \R}$ for any
$s,o \in \setent$ and $p \in \setrel$ which predicts the truth value of possible triples. We will write $\Ne = |\setent|$ and $\Nr = |\setrel|$.

\noindent
For notational convenience, we define the trilinear product of three complex
vectors as:
\begin{equation*}
   \langle a,b,c \rangle = \sum\limits_{j=1}^\rk a_j b_j c_j = a\T (b \odot c)
\end{equation*}
where $a,b,c \in \C^\rk$, and $\odot$
denotes the Hadamard product, i.e. the element-wise
product between two vectors of same length.

\noindent
In the following, we will consider the discrete Fourier transform (DFT)
of purely real vectors only : ${\mathcal{F}: \R^\rk \rightarrow \C^\rk}$.
For $j \in \{0,\ldots,\rk-1\}$:
\begin{equation}
  \F(x)_j = \sum_{k=0}^{\rk-1} x_k e^{-2i \pi j \frac{k}{\rk}}
  \label{eq:dft}
\end{equation}
where $\F(x)_j \in \C$ is the $j^{\mathrm{th}}$ value in the resulting complex
vector $\F(x) \in \C^\rk$. 
Note that the components in \Cref{eq:dft} are indexed
from 0 to $K-1$.

\subsection{Holographic Embeddings}

The holographic embeddings model (\textsc{HolE}) \cite{nickel_2016_holographic} represents relations and
entities with real-valued embeddings 
$\Em \in \R^{\Ne \times \rk}$,
$\Rm \in \R^{\Nr \times \rk}$, 
and scores a triple $(p,s,o)$ with 
the dot product between the embedding of the relation $p$ and
the circular correlation $\star: \R^\rk \times \R^\rk \rightarrow \R^\rk$ of the embeddings of entities
$s$ and $o$:
\begin{equation}
    \score^h_{pso} = \rv_p\T ( \ev_s \star \ev_o)\,.
    \label{eq:hole}
\end{equation}
The circular correlation can be written with the
discrete Fourier transform (DFT), 
\begin{equation}
     \ev_s \star \ev_o = \F^{-1}(\overline{\F(\ev_s)} \odot \F(\ev_o))
    \label{eq:ccorr_to_fourier}
\end{equation}
where $\mathcal{F}^{-1}: \C^\rk \rightarrow \C^\rk$ is the inverse DFT.
In this case, the embedding vectors are
real $\ev_s, \ev_o, \rv_p \in \R^\rk$, and so is the result
of the inverse DFT, since the circular correlation
of real-valued vectors results in a real-valued vector.

\subsection{Complex Embeddings}

The complex embeddings model (\textsc{ComplEx}) \cite{trouillon2016,trouillon2017knowledge} represents relations and
entities with complex-valued embeddings 
$\Em \in \C^{\Ne \times \rk}$,
$\Rm \in \C^{\Nr \times \rk}$, 
and scores a triple $(p,s,o)$ with 
the real part of the trilinear product of the corresponding
embeddings:
\begin{equation}
    \score^c_{pso} = \real\left(\langle \rv_p,\ev_s,\conj{\ev}_o\rangle\right)
    \label{eq:complex}
\end{equation}
where $\ev_s, \ev_o, \rv_p \in \C^\rk$ are complex vectors,
and $\conj{\ev}_o$ is the complex conjugate of the vector $\ev_o$.

\subsection{Equivalence}
The equivalence of \textsc{HolE} and \textsc{ComplEx} has recently been shown by \citet{hayashi2017equivalence}.
In the following, we briefly discuss this equivalence of both models and how it can be derived. For completeness, a full proof similar to that of \citet{hayashi2017equivalence} is included in \Cref{sec:equiv-proof}.

First, to derive the connection between \textsc{HolE} and \textsc{ComplEx}, consider
\textit{Parseval's Theorem}:
\begin{thm}
    Suppose $x,y \in \R^\rk$ are real vectors. Then $x\T y = \frac{1}{\rk}
    \F(x)\T \conj{\F(y)} $.
    \label{thm:parseval}
\end{thm}

\noindent
Using \Cref{thm:parseval} as well as \Cref{eq:hole,eq:ccorr_to_fourier}, we can
then rewrite the scoring function of \textsc{HolE} as:
\begin{align}
    \score^h(p,s,o) &= \rv_p\T ( \ev_s \star \ev_o)\label{eq:hole-fft}\\
     &= \rv_p\T (\F^{-1}(\overline{\F(\ev_s)} \odot \F(\ev_o)))\notag\\
     &= \frac{1}{\rk} \F(\rv_p)\T \conj{\F(\F^{-1}(\overline{\F(\ev_s)} \odot \F(\ev_o)))}\notag\\
     &= \frac{1}{\rk} \F(\rv_p)\T (\F(\ev_s) \odot \conj{\F(\ev_o)})\notag\\
     &= \frac{1}{\rk} \left< \F(\rv_p), \F(\ev_s), \conj{\F(\ev_o)} \right>\,.
     \label{eq:corr_to_fourier}
\end{align}
Hence, for \textsc{HolE} we could directly learn \emph{complex} embeddings
$\widehat{\ev}_i \equiv \F(\ev_i), \widehat{\rv}_j \equiv \F(\rv_j) \in \mathbb{C}^d$ instead of learning embeddings $\ev_i,
\rv_j \in \R^d$ and mapping them into the frequency domain and back. However, to
ensure that the trilinear product of these complex embeddings is a real number,
we would either need to enforce the same symmetry constraints on $F(\ev_i)$ and
$F(\rv_j)$ that arise from the DFTs or---alternatively---take only the
real-valued part of the trilinear product. We show in \Cref{sec:equiv-proof} that these are two ways of performing
the same operation, hence showing that the scoring functions of
\textsc{ComplEx} and \textsc{HolE} are equivalent---up to a
constant factor.

Furthermore, both models have equal memory complexity,
as the equivalent complex vectors are twice as small 
(see proof in \Cref{sec:equiv-proof})
but require twice as much memory as real-valued ones of same size---for
a given floating-point precision.
However, the complex formulation of the scoring function 
reduces the time complexity
from $\mathcal{O}(\rk \log(\rk))$ (quasilinear) to $\mathcal{O}(\rk)$ (linear). 

\section{Loss Functions \& Predictive Abilities}
\label{sec:exp-loss}

The experimental results of \textsc{HolE} and \textsc{ComplEx} as reported by
\citet{nickel_2016_holographic} and \citet{trouillon2016}
agreed on the WN18 data set,
but diverged significantly on FB15K \cite{bordes2014semantic}---although both scoring function are equivalent.
Since the main difference in the experimental settings was the use of different loss functions---i.e., margin loss versus logistic loss---we analyze in this section whether the discrepancy of results can be attributed to this fact.
For this purpose, we implemented both loss functions for the complex
representation $\phi^c$ within the same framework,
and compared the results on the WN18 and FB15K data sets. 

First, note that in both data sets, only positive training triples
are provided. Negative examples are generated by corrupting 
the subject or object entity of each positive triple, as 
described in \citet{bordes2013translating}.
In the original \textsc{HolE} publication \cite{nickel_2016_holographic}, 
a pairwise margin
loss is optimized over each positive and its corrupted
negative $(p,s',o')$:
\begin{equation}
    \mathcal{L}(\setobs;\Theta) =
    \sum_{\mathclap{((p,s,o),y) \in \setobs}} [\gamma + \sigma(\phi^h_{ps'o'}) - \sigma(\phi^h_{pso})]_+
    \label{eq:ll_objective}
\end{equation}
where $\gamma$ is the margin hyperparameter, and $\sigma$ the standard logistic function. The entity embeddings are also constrained to unit norm : 
$||e_i||_2 \leq 1$, for all $i \in \setent$.

Whereas in \citet{trouillon2016}, the generated negatives
are merged into the training set $\setobs$ at each batch sampling,
and the log-likelihood is optimized with $L^2$ regularization:
\begin{equation}
    \mathcal{L}(\setobs;\Theta) =
    \sum_{\mathclap{((p,s,o),y) \in \setobs}} \log( 1 + \exp(-y\phi^c_{pso})) + \lambda ||\Theta||^2_2 \,.
    \label{eq:margin_objective}
\end{equation}

Optimization is conducted
with stochastic gradient descent, AdaGrad \cite{duchi2011adaptive},
and early stopping, as described in
\citet{trouillon2016}. A single
corrupted negative triple is generated for each positive training triple.
The results are reported for the best validated models after grid-search on the
following values: $\rk \in \{$10, 20, 50, 100, 150, 200$\}$, $\lambda \in
\{$0.1, 0.03, 0.01, 0.003, 0.001, 0.0003, 0.0$\}$ for the log-likelihood loss, and
$\gamma \in \{$0.1, 0.2, 0.3, 0.4, 0.5, 0.6, 0.7, 0.8, 0.9, 1.0$\}$ for the max-margin loss.
The raw and filtered mean reciprocal ranks (MRR),
as well as the filtered hits at 1, 3 and 10 are reported in Table 
\ref{tab:fb15k_wn18_res}.

\begin{table*}[t]
    \centering
    \resizebox{\textwidth}{!}{%
    \begin{tabular}{@{\extracolsep{8pt}}lllllllllll@{}}
        \toprule
        
         & \multicolumn{5}{c}{\textbf{WN18}} & \multicolumn{5}{c}{\textbf{FB15K}} \\ \cmidrule(lr){2-6} \cmidrule(l){7-11}
         & \multicolumn{2}{c}{MRR} & \multicolumn{3}{c}{Hits at} & \multicolumn{2}{c}{MRR} & \multicolumn{3}{c}{Hits at} \\ \cmidrule(lr){2-3} \cmidrule(lr){4-6} \cmidrule(lr){7-8} \cmidrule(l){9-11}
        
        Loss & Filtered & Raw & 1 & 3 & 10 & Filtered & Raw & 1 & 3 & 10 \\ \midrule
        Margin & 0.938 & \textbf{0.605} & 0.932 & 0.942 & \textbf{0.949} & 0.541 & \textbf{0.298} & 0.411 & 0.627 & 0.757\\
        Neg-LL & \textbf{0.941} &  0.587 &  \textbf{0.936} &  \textbf{0.945} &  0.947 & \textbf{0.639} & 0.250 & \textbf{0.523} & \textbf{0.725} & \textbf{0.825}\\
        \bottomrule
    \end{tabular}
    }
    \caption{Filtered and raw mean reciprocal rank (MRR), Hits@N metrics are filtered, for the pairwise max-margin loss and the negative log-likelihood on WN18 and FB15K data sets.}
    \label{tab:fb15k_wn18_res}
\end{table*}

The margin loss results are consistent with the \textsc{HolE}
ones originally reported
in \citet{nickel_2016_holographic},
which confirms the equivalence of the scoring functions, 
and supports the hypothesis that the loss was responsible for the 
difference in previously reported results.
The log-likelihood results are also coherent, as one must 
note that the higher scores reported on FB15K in \citet{trouillon2016}
are due to the use of more than one generated negative sample
for each positive training triple. Here, we generated a single negative
sample for each positive one in order to keep the
comparison fair between the two losses.
The max-margin loss achieves a better raw MRR (rankings 
without removing the training samples) on both datasets, 
but much worse filtered metrics on FB15K, suggesting that 
this loss can be more prone to overfitting.

\section{Scoring Function \& Symmetry}

The results in \Cref{sec:exp-loss} suggest that the choice of scoring
function, i.e., \textsc{ComplEx} or \textsc{HolE}, does not affect the
predictive abilities of the model.
An additional important question is whether one of the models---in practice---is
better suited for modeling certain types of relations. In particular, for
symmetric relations, \textsc{HolE} needs to learn embeddings for which the
imaginary part \emph{after} the DFT is close to zero. \textsc{ComplEx}, on the other hand,
can learn such representations easily as it operates directly in the complex
domain.
The question whether this difference in models translates to differences in
practice affects the learning of both symmetric and antisymmetric
relations. Relations $p \in \setrel$ are symmetric when triples have the same truth value by permutation of the subject and object entities: $y_{pso} = y_{pos}$ for all $s,o \in \setent$, whereas
facts of antisymmetric relations $p$ have inverse truth values:  $y_{pso} = - y_{pos}$.
To evaluate this question
experimentally, we reproduced the joint learning of synthetic symmetric and antisymmetric
relations described in \citet{trouillon2016} on both scoring functions.
We used the log-likelihood loss as all negatives are observed.

\begin{figure*}[t]{{\extracolsep{8pt}}}
	\centering
	\includegraphics[width=0.298\textwidth]{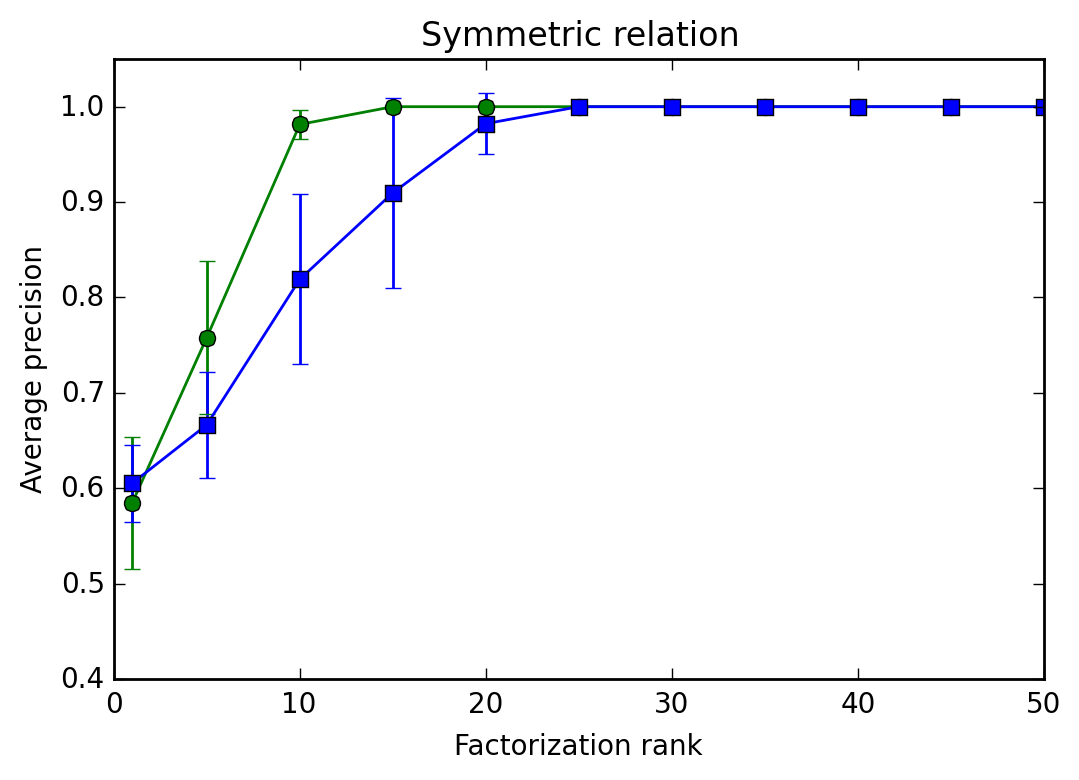}
	\includegraphics[width=0.298\textwidth]{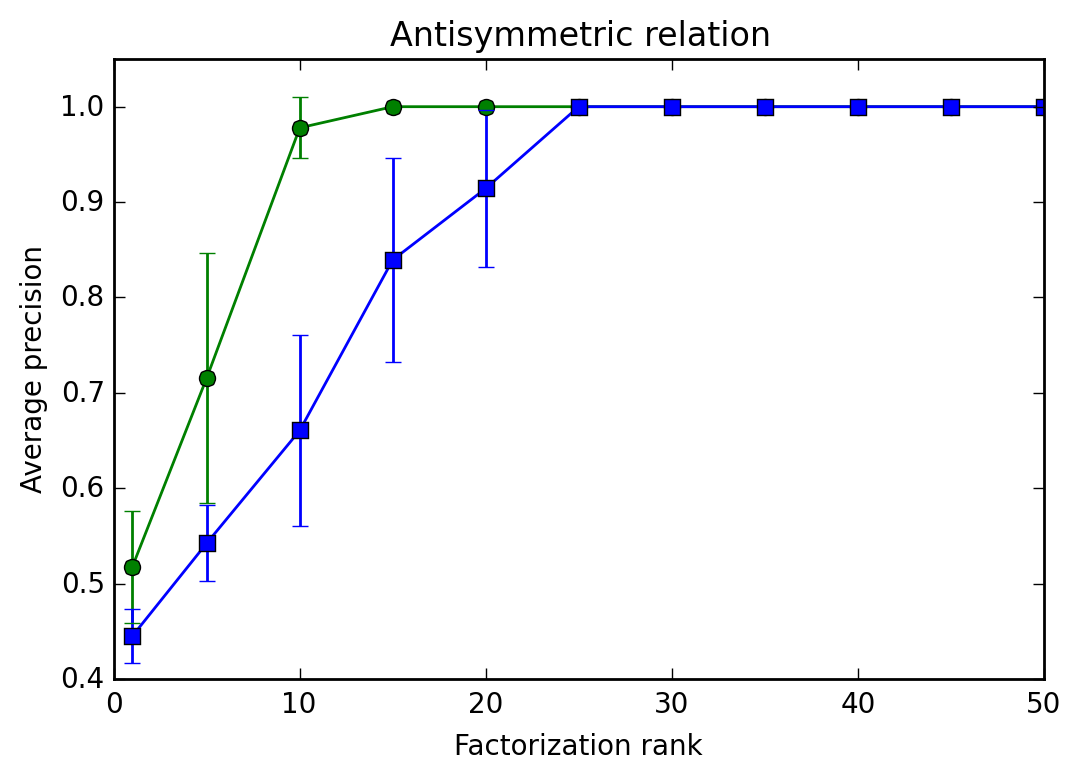} 
	\includegraphics[width=0.393\textwidth]{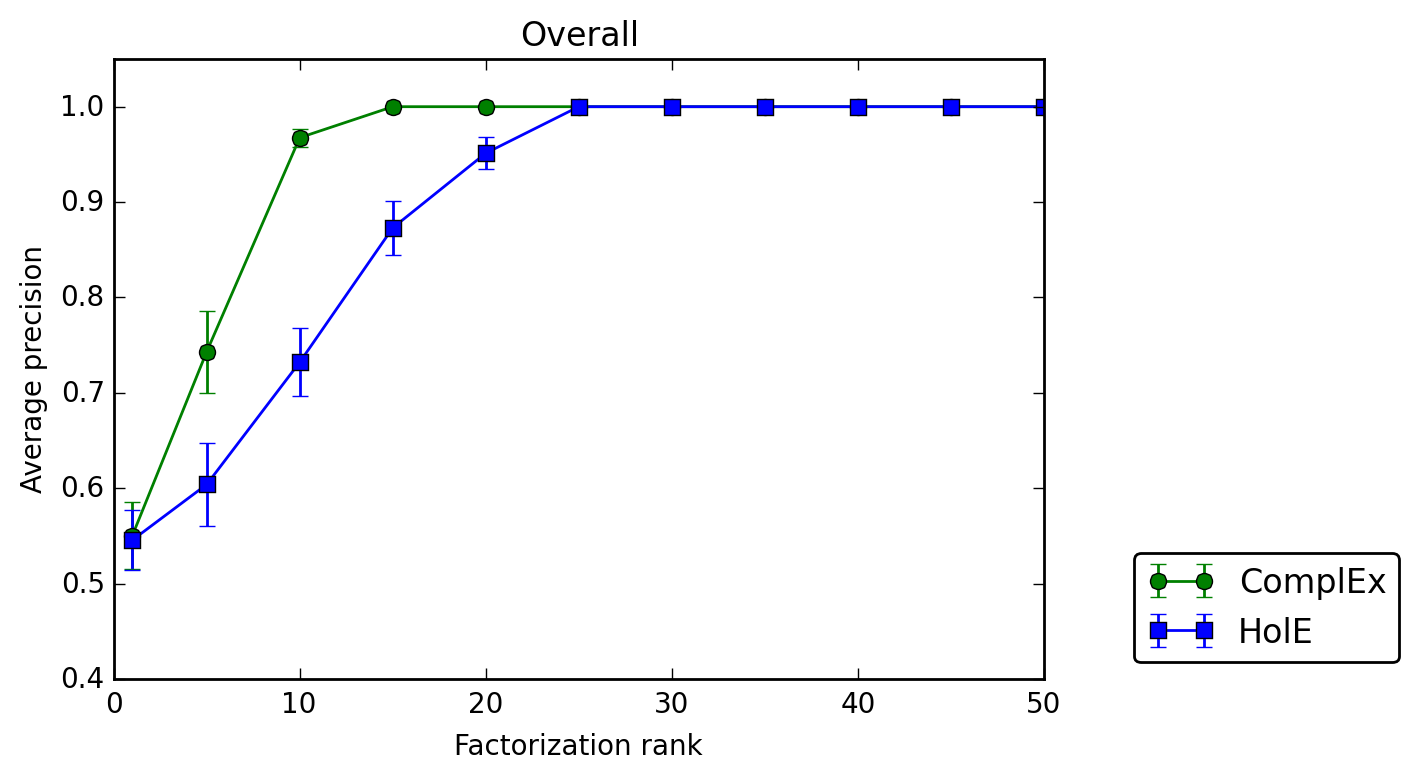}
	\caption{Average precision (AP) for each factorization rank ranging from 1 to 50 for
	the HolE and ComplEx scoring functions, with log-likelihood loss. Learning is performed jointly on the symmetric relation and on the antisymmetric relation. Top-left: AP over the symmetric relation only. Top-right: AP over the antisymmetric relation only. Bottom: Overall AP.}
	\label{fig:exp_sym_antisym}
\end{figure*}

We generated randomly a $50 \times 50$ symmetric matrix, and 
a $50 \times 50$ antisymmetric matrix. Jointly, they represent
a $2 \times 50 \times 50$ tensor. To ensure that all test values are predictable, the upper triangular parts of the matrices are always kept in the training set, and the diagonals are unobserved. We conducted 5-fold cross-validation on the lower-triangular matrices, using the upper-triangular parts plus 3 folds for training, one fold for validation and one fold for testing. The regularization parameter $\lambda$ is validated among the same values as in the previous experiment.

Figure \ref{fig:exp_sym_antisym} shows the best cross-validated average precision (area under the precision-recall curve) for the two scoring functions 
for ranks ranging up to 50.
Both models manage to perfectly model symmetry and antisymmetry.
As the ComplEx model has twice has many parameters for a given rank, 
it reaches a perfect
average precision with a twice smaller rank. This confirms that 
the representation of the scoring function does not affect the learning
abilities of the models in practice.

\section{Discussion}

We have demonstrated that the scoring functions of the \textsc{HolE} and
\textsc{ComplEx} models are directly proportional. 
This hence extends the existence property of
the \textsc{ComplEx} model over all knowledge graphs \cite{trouillon2017knowledge}
to the \textsc{HolE} model.
We also showed experimentally
that the difference between the reported results of the two models was due to
the use of different loss functions, and specifically that the log-likelihood
loss can produce a large improvement of predictive performances over the more
often used margin loss. We have also shown that Complex and Holographic
embeddings can be trained equally well on symmetric and antisymmetric patterns.
All these things being equal, an interesting question is then in which settings one of the two models is
preferable. Complex embeddings have an advantage in terms of time complexity as
they scale linearly with the embedding dimension, whereas Holographic embeddings
scale quasilinearly. An advantage of Holographic embeddings however is that the
embeddings remain strictly in the real domain, which makes it easier for them to
be used in other real-valued machine learning models. 
In contrast, Complex embeddings can not
easily be transformed to real-valued vectors and used without
loss of information---i.e. the specific way the 
real and imaginary parts interact in algebraic operations.
Complex-valued models in which Complex embeddings can be directly
input are emerging in machine
learning \cite{trabelsi2017deep,danihelka2016associative}, but
this path is yet to be explored for 
other relational learning problems.
Hence,
if the task of interest is link prediction, Complex embeddings offer an improved runtime complexity in the order of $O(\log \rk)$. If the embeddings should
be used in further machine learning models, e.g. for entity classification,
Holographic embeddings provide better compatibility with existing real-valued methods.

Furthermore, while the choice of the loss is of little consequence on
the WN18 dataset, our experiments showed that the log-likelihood loss performed significantly better on FB15K. While much research attention has been
given to scoring functions in link prediction, little has been said
about the losses, and the max-margin loss has been used
in most of the existing work
\cite{bordes2013translating,Yang2015,riedel_2013_univschema}. An interesting direction of future work is therefore a more detailed study of loss functions for knowledge graph embeddings---especially in light of the highly skewed label distribution and the open-world assumption which are characteristic for knowledge graphs but unusual for standard machine learning settings.

\section*{Acknowledgments}
This work was supported in part by the Association Nationale de la
Recherche et de la Technologie through the CIFRE grant 2014/0121.

\appendix

\section{Proof of Equivalence}
\label{sec:equiv-proof}
In this section, we provide the full proof for the equivalence of both models. 
Note that a similar proof has recently been derived by \citet{hayashi2017equivalence}.

We start from \Cref{eq:hole-fft} and show that there always
exists corresponding real-valued holographic embeddings
and complex embeddings such that the scoring functions of
\textsc{HolE} and \textsc{ComplEx} are directly proportional, i.e.
they are mathematically equal up to a constant multiplier $a \in \R$:
$\score^h_{pso} = a \score^c_{pso}$.
The key idea is in showing that the symmetry structure
of vectors resulting from Fourier transform of real-valued vectors
is such that, the trilinear product between these structured vectors
is actually equal to keeping the real part
of the trilinear product of their first half.

First, we derive a property of the DFT on real vectors $x$,
showing that the resulting complex vector $\F(x)$ has
a partially symmetric structure, for $j \in \{1,\ldots,\rk-1\}$:
\begin{align}
    \F(x)_{(\rk-j)} &= \sum_{k=0}^{\rk-1} x_k e^{-2i \pi (K-j) \frac{k}{\rk}}\notag\\
    &= \sum_{k=0}^{\rk-1} x_k e^{-2i \pi k} e^{2i \pi j \frac{k}{\rk}}\notag\\
    \intertext{and given that $k$ is an integer: $e^{-2i \pi k} = 1$,}
    &= \sum_{k=0}^{\rk-1} x_k e^{2i \pi j \frac{k}{\rk}}
    = \sum_{k=0}^{\rk-1} x_k \conj{e^{-2i \pi j \frac{k}{\rk}}}\notag\\
    \intertext{and since $x_k \in \R$,}
    &= \sum_{k=0}^{\rk-1} \conj{x_k e^{-2i \pi j \frac{k}{\rk}}}
    = \conj{\F(x)_j}\,.
    \label{eq:sym_prop}
\end{align}

Two special cases arise, the first one is $F(x)_0$, which is not concerned
by the above symmetry property:
\begin{align}
    \F(x)_0 &= \sum_{k=0}^{\rk-1} x_k e^{-2i \pi 0 \frac{k}{\rk}}\notag\\
    &= \sum_{k=0}^{\rk-1} x_k
    =: s(x) \in \R\,.
                                  \notag\\
    \label{eq:f_0}
\end{align}
And the second one is $F(x)_{\frac{\rk}{2}}$ when $K$ is even:
\begin{align}
    \F(x)_{(\rk-\frac{\rk}{2})} & = \conj{ \F(x)_{\frac{\rk}{2}}} =  \F(x)_{\frac{\rk}{2}}\notag\\
    &= \sum_{k=0}^{\rk-1} x_k e^{-2i \pi \frac{\rk k}{2 \rk}} 
    = \sum_{k=0}^{\rk-1} x_k e^{-i \pi k} \notag\\
                                &= \sum_{k=0}^{\frac{\rk}{2}-1} x_{2k} - x_{2k + 1}
    =: t(x) \in \R\,.
                                  \notag\\
    \label{eq:f_k/2}
\end{align}

From \Cref{eq:sym_prop,eq:f_0,eq:f_k/2}, we write
the general form of the Fourier transform $\F(x) \in \C^K$ of a real vector $x \in \R^K$:
\begin{equation}
\F(x)=
\begin{cases}
    [s(x) \enspace x' \enspace t(x) \enspace\conj{x'}_{\leftarrow}], & \text{if}\ K\ \text{is even,} \\
    [s(x) \enspace x' \enspace\conj{x'}_{\leftarrow}], & \text{if}\ K\ \text{is odd.}
\end{cases}
\end{equation}
where $x', x'_{\leftarrow} \in \C^{\left\lceil \frac{\rk}{2} \right\rceil - 1}$, 
with $x' = [ \F(x)_1, \ldots , \F(x)_{\left\lceil \frac{\rk}{2} \right\rceil -1} ]$, 
and $x'_{\leftarrow}$ is $x'$ in reversed order: 
$x'_{\leftarrow} = [ \F(x)_{\left\lceil \frac{\rk}{2} \right\rceil -1}, \ldots , \F(x)_1 ]$.

We can then derive \Cref{eq:corr_to_fourier} for $\rv_p,\ev_s,\ev_o \in \R^\rk$, first with $K$ being odd:
\begin{align}
\score^h_{pso} &= \frac{1}{\rk} \left< \F(\rv_p), \F(\ev_s), \conj{\F(\ev_o)} \right>\notag\\
   &= \frac{1}{\rk}  \left<[s(\rv_p) \enspace \rv'_p\enspace\conj{\rv}'_{p\leftarrow}],  [s(\ev_s) \enspace \ev'_s\enspace\conj{\ev}'_{s\leftarrow}], \conj{[s(\ev_o) \enspace \ev'_o\enspace\conj{\ev}'_{o\leftarrow}]}\right>\notag\\
   &= \frac{1}{\rk}  \left<[s(\rv_p) \enspace \rv'_p\enspace\conj{\rv}'_{p}], [s(\ev_s) \enspace \ev'_s\enspace\conj{\ev}'_{s}], [s(\ev_o) \enspace \conj{\ev}'_o\enspace\ev'_{o}]\right>\notag\\
   &= \frac{1}{\rk} \left( s(\rv_p) s(\ev_s) s(\ev_o) + \left<\rv'_p,\ev'_s,\conj{\ev}'_o\right> + \left<\conj{\rv}'_p,\conj{\ev}'_s,\ev'_o \right> \right) \notag\\
   &= \frac{1}{\rk} \left(  s(\rv_p) s(\ev_s) s(\ev_o) + \left<\rv'_p,\ev'_s,\conj{\ev}'_o \right> + \conj{\left<\rv'_p,\ev'_s,\conj{\ev}'_o\right>} \right) \notag\\
   &= \frac{1}{\rk} \left(  s(\rv_p) s(\ev_s) s(\ev_o) + 2\, \real\left({\left<\rv'_p,\ev'_s,\conj{\ev}'_o\right>}\right) \right) \notag\\
   &= \frac{2}{\rk}  \real\left(\left<\left[\sqrt[3]{\tfrac{1}{2}} s(\rv_p) \enspace \rv'_p\right], \left[\sqrt[3]{\tfrac{1}{2}} s(\ev_s) \enspace \ev'_s\right], \left[\sqrt[3]{\tfrac{1}{2}} s(\ev_o) \enspace \conj{\ev}'_o\right]\right>\right)\notag\\
   &= \frac{2}{\rk}  \real\left({\left<\rv''_p,\ev''_s,\conj{\ev}''_o\right>}\right)\notag\\
   &= \frac{2}{\rk} \score^c_{pso} \,
\end{align}
where $\rv''_p,\ev''_s,\ev''_o \in \C^{\left\lceil \frac{\rk}{2} \right\rceil}$.
The derivation is similar when $K$ is even, with double prime vectors being $x'' = [\sqrt[3]{\tfrac{1}{2}} s(x) \enspace \sqrt[3]{\tfrac{1}{2}} t(x) \enspace x'] \in \C^{\frac{\rk}{2} + 1}$.

As mentioned in \Cref{sec:equiv}, the complex vectors $\rv''_r,\ev''_s,\ev''_o \in \C^{\left\lceil \frac{\rk}{2} \right\rceil}$  equivalent to the real vectors 
$\rv_p,\ev_s,\ev_o \in \R^\rk$ are twice smaller,
but take twice as much memory as real-valued ones of same size at
a given floating-point precision. Both models
hence have the exact same memory complexity.

\bibliography{complex_refs}
\bibliographystyle{aaai}

\end{document}